\documentclass[11pt,letterpaper,pdftex]{article}
\usepackage{ijcnlp2017}
\usepackage{times}
\usepackage{latexsym}

\usepackage{avm}
\usepackage[pdftex]{graphicx}
\usepackage{amsmath, amssymb}
\usepackage{booktabs}

\usepackage{color}

\ijcnlpfinalcopy

\title{Revisiting the Design Issues of Local Models for \\ Japanese Predicate-Argument Structure Analysis}

\author{
Yuichiroh Matsubayashi$^\spadesuit$ \and Kentaro Inui$^{\spadesuit\diamondsuit}$ \\
{$^\spadesuit$Tohoku University, $^\diamondsuit$RIKEN Center for Advanced Intelligence Project}\\
{\tt \{y-matsu, inui\}@ecei.tohoku.ac.jp}
}

\date{}

\begin{document}

\maketitle

\begin{abstract}

The research trend in Japanese predicate-argument structure (PAS) analysis is shifting from pointwise prediction models with local features to global models designed to search for globally optimal solutions.
However, the existing global models tend to employ only relatively simple local features; therefore, the overall performance gains are rather limited.
The importance of designing a local model is demonstrated in this study by showing that the performance of a sophisticated local model can be considerably improved with recent feature embedding methods and a feature combination learning based on a neural network, outperforming the state-of-the-art global models in $F_1$ on a common benchmark dataset.
\end{abstract}

\section{Introduction}
A predicate-argument structure (PAS) analysis is the task of analyzing the structural relations between a predicate and its arguments in a text and is considered as a useful sub-process for a wide range of natural language processing applications~\cite{Shen2007,DBLP:conf/acl/KudoIK14,Liu2015a}.

PAS analysis can be decomposed into a set of primitive subtasks that seek a filler token for each argument slot of each predicate.
The existing models for PAS analysis fall into two types: local models and global models. Local models independently solve each primitive subtask in the pointwise fashion~\cite{Seki2002,taira2008japanese,imamura2009discriminative,Yoshino2013}.
Such models tend to be easy to implement and faster compared with global models but cannot handle dependencies between primitive subtasks.
Recently, the research trend is shifting toward global models that search for a globally optimal solution for a given set of subtasks by extending those local models with an additional ranker or classifier that accounts for dependencies between subtasks~\cite{Iida2007,komachi2010,yoshikawa2011jointly,hayashibe2014,Ouchi2015,Iida2015,Iida2016,Shibata2016a}.

However, even with the latest state-of-the-art global models~\cite{Ouchi2015,ouchi2017},
the best achieved $F_1$ remains as low as $81.4\%$ on a commonly used benchmark dataset~\cite{Iida2007ntc},
wherein the gain from the global scoring is only 0.3 to 1.0 point.
We speculate that one reason for this slow advance is that recent studies pay too much attention to global models and thus tend to employ overly simple feature sets for their base local models.

The goal of this study is to argue the importance of designing a sophisticated local model before exploring global solution algorithms and to demonstrate its impact on the overall performance through an extensive empirical evaluation.
In this evaluation, we show that a local model alone is able to significantly outperform
the state-of-the-art global models by incorporating a broad range of local features
proposed in the literature and training a neural network for combining them.
Our best local model achieved $13$\% error reduction in $F_1$ compared with the state of the art.

\section{Task and Dataset}
\label{sec:task}
In this study, we adopt the specifications of the NAIST Text Corpus (NTC)~\cite{Iida2007ntc}, a commonly used benchmark corpus annotated with nominative (NOM), accusative (ACC), and dative (DAT) arguments for predicates.
Given an input text and the predicate positions, the aim of the PAS analysis is to identify the head of the filler tokens for each argument slot of each predicate.

The difficulty of finding an argument tends to differ depending on the relative position of the argument filler and the predicate.
In particular, if the argument is omitted and the corresponding filler appears outside the sentence, the task is much more difficult because
we cannot use the syntactic relationship between the predicate and the filler in a naive way.
For this reason, a large part of previous work narrowed the focus to the analysis of arguments in a target sentence \cite{yoshikawa2011jointly, Ouchi2015, Iida2015}, and here, we followed this setting as well.

\section{Model}
\label{sec:model}

Given a tokenized sentence $s$ and a target predicate $p$ in $s$ with the gold dependency tree $t$,
the goal of our task is to select at most one argument token $\hat{a}$ for each case slot of the target predicate.

\begin{figure}[t!]
  \centering
	\includegraphics[width=1.0\linewidth]{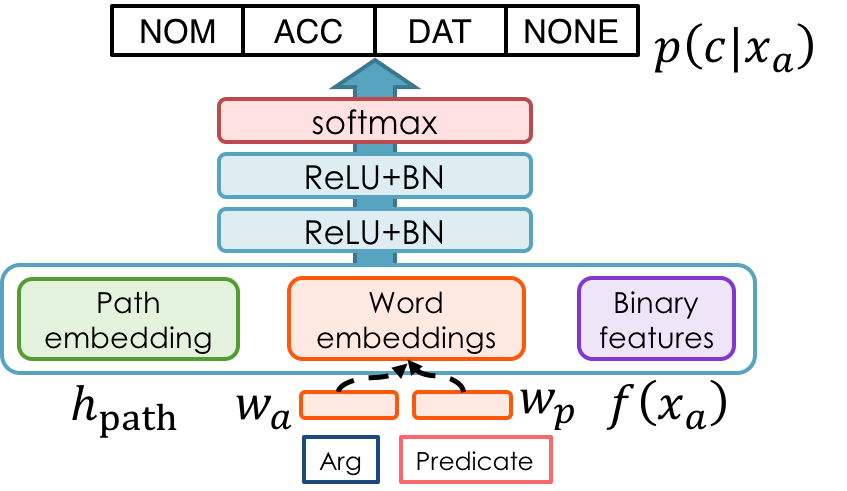}
  \caption{Network structure of our NN model}
	\label{fig:network}
\end{figure}

Taking $x_a = (a,p,s,t)$ as input, our model estimates the probability $p(c|x_a)$ of assigning a case label $c \in \{ \mathtt{NOM}, \mathtt{ACC}, \mathtt{DAT}, \mathtt{NONE} \}$ for each token $a$ in the sentence, and then selects a token with a maximum probability that exceeds the output threshold $\theta_c$ for $c$.
The probability $p(c|x_a)$ is modeled by a neural network (NN) architecture, which is a fully connected multilayer feedforward network stacked with a softmax layer on the top (Figure~\ref{fig:network}).
\begin{eqnarray}
	g &=& \mathrm{softmax}(W_{n+1}h_n + b_{n+1}) \\
	h_i &=& \mathrm{ReLU}(\mathrm{BN}(W_i h_{i-1} + b_i)) \\
	h_1 &=& \mathrm{ReLU}(\mathrm{BN}(W_1 m + b_1)) \\
  m &=& [h_{\rm path}, w_p, w_a, f(x_a)] \label{eq:input-layer}
\end{eqnarray}

The network outputs the probabilities $g$ of assigning each case label for an input token $a$, from automatically learned combinations of feature representations in input $m$.
Here, $h_i$ is an $i$-th hidden layer and $n$ is the number of hidden layers. We apply batch normalization (BN) and a ReLU activation function to each hidden layer.

The input layer $m$ for the feedforward network is a concatenation of the three types of feature representations described below: a path embedding $h_{\rm path}$, word embeddings of the predicate and the argument candidate $w_p$ and $w_a$, and a traditional binary representation of other features $f(x_a)$.

\subsection{Lexicalized path embeddings}
\label{ssec:path-emb}

\begin{figure}[t]
  \centering
	\includegraphics[width=1.0\linewidth]{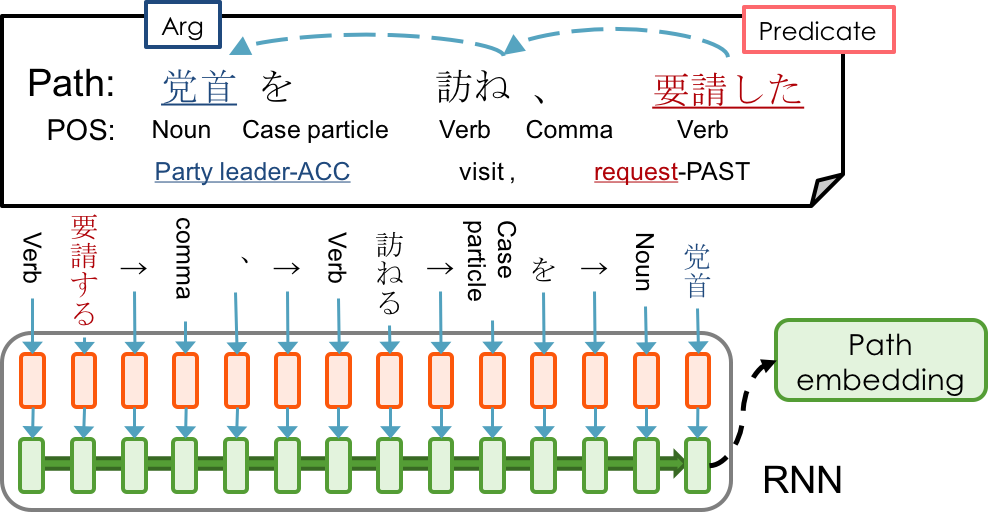}
	\caption{Path embedding}
	\label{fig:path-embedding}
\end{figure}

When an argument is not a direct dependent of a predicate, the dependency path is considered as important information.
Moreover, in some constructions such as raising, control, and coordination, lexical information of intermediate nodes is also beneficial although a sparsity problem occurs with a conventional binary encoding of lexicalized paths.

\newcite{Roth2016a} and \newcite{Shwartz2016}
recently proposed methods for embedding a lexicalized version of dependency path on a single vector using RNN.
Both the methods embed words, parts-of-speech, and directions and labels of dependency in the path into a hidden unit of LSTM and output the final state of the hidden unit.
We adopt these methods for Japanese PAS analysis and compare their performances.

As shown in Figure~\ref{fig:path-embedding}, given a dependency path from a predicate to an argument candidate, we first create a sequence of POS, lemma, and dependency direction for each token in this order by traversing the path.\footnote{
We could not use dependency labels in the path since traditional parsing framework in Japanese does not have dependency labels. However, particles in Japanese can roughly be seen as dependency relationship markers, and, therefore, we think these adaptations approximate the original methods.
}
Next, an embedding layer transforms the elements of this sequence into vector representations.
The resulting vectors are sequentially input to RNN. Then, we use the final hidden state as the path-embedding vector.
We employ GRU \cite{Cho2014gru} for our RNN and use two types of input vectors:
the adaptations of \newcite{Roth2016a}, which we described in Figure~\ref{fig:path-embedding},
and \newcite{Shwartz2016}, which concatenates vectors of POS, lemma and dependency direction for each token into a single vector.

\subsection{Word embedding}
\label{ssec:word-emb}
The generalization of a word representation is one of the major issues in SRL.
\newcite{Fitzgerald2015} and \newcite{Shibata2016a} successfully improved the classification accuracy of SRL tasks by generalizing words using embedding techniques. We employ the same approach as \newcite{Shibata2016a}, which uses the concatenation of the embedding vectors of a predicate and an argument candidate.

\begin{table}[!t]
  \centering
	\footnotesize{
	\begin{tabular}{l|l}
		\toprule
For  & surface, lemma, POS, \\
predicate & type of conjugated form, \\
$p$   	        & nominal form of nominal verb, \\
        & voice suffixes (-reru, -seru, -dekiru, -tearu)\\
\midrule
For    & surface, lemma, POS, NE tag, \\
argument  & whether $a$ is head of {\it bunsetsu}, \\
candidate & particles in {\it bunsetsu}, \\
$a$ & right neighbor token's lemma and POS \\
\midrule
Between & case markers of other dependents of $p$, \\
predicate & whether $a$ precedes $p$, \\
and & whether $a$ and $p$ are in the same {\it bunsetsu}, \\
argument & token- and dependency-based distances, \\
candidate & naive dependency path sequence \\
    \bottomrule
	\end{tabular}
	}
	\caption{Binary features}
	\label{tbl:binary-features}
\end{table}

\subsection{Binary features}
\label{ssec:binary-features}
\paragraph{\bf Case markers of the other dependents}
Our model independently estimates label scores for each argument candidate.
However, as argued by \newcite{toutanova2008global} and \newcite{yoshikawa2011jointly}, there is a dependency between the argument labels of a predicate.

In Japanese, case markers (case particles) partially represent a semantic relationship between words in direct dependency.
We thus introduce a new feature that approximates co-occurrence bias of argument labels by gathering case particles for the other direct dependents of a target predicate.

\paragraph{\bf Other binary features}

The other binary features employed in our models have mostly been discussed in
previous work~\cite{imamura2009discriminative,Hayashibe2011}.
The entire list of our binary features are presented in Table~\ref{tbl:binary-features}.

\section{Experiments}
\label{sec:experiments}

\begin{table*}[!t]
  \centering
	\footnotesize{
	\begin{tabular}{l|l||l||r|r|r|r|r|r|r|r}
		\toprule
		
    & & \multicolumn{3}{c|}{All} & \multicolumn{6}{c}{$F_1$ in different dependency distance} \\
		Model & Binary feats. & $F_1$ ($\sigma$) & Prec. & Rec. & {\it Dep} & {\it Zero} & 2 & 3 & 4 & $\geq$ 5 \\
		\midrule
    B               & all                  & 82.02 ($\pm$0.13) & 83.45 & 80.64 & 89.11 & 49.59 & 57.97 & 47.2 & 37  & 21  \\
    B               & $-$cases             & 81.64 ($\pm$0.19) & 83.88 & 79.52 & 88.77 & 48.04 & 56.60 & 45.0 & 36  & 21  \\
    \midrule
    WB              & all                  & 82.40 ($\pm$0.20) & 85.30 & 79.70 & 89.26 & 49.93 & 58.14 & 47.4 & 36  & 23  \\
    WBP-Roth        & all                  & 82.43 ($\pm$0.15) & 84.87 & 80.13 & 89.46 & 50.89 & 58.63 & 49.4 & 39  & {\bf 24}  \\
    WBP-Shwartz     & all                  & 83.26 ($\pm$0.13) & 85.51 & 81.13 & 89.88 & 51.86 & 60.29 & 49.0 & 39  & 22  \\
    WBP-Shwartz     &  $-$word             & 83.23 ($\pm$0.11) & 85.77 & 80.84 & 89.82 & 51.76 & 60.33 & 49.3 & 38  & 21  \\
    WBP-Shwartz     &  $-$\{word, path\}   & 83.28 ($\pm$0.16) & 85.77 & 80.93 & 89.89 & 51.79 & 60.17 & 49.4 & 38  & 23  \\
    WBP-Shwartz (ens) &  $-$\{word, path\}    & {\bf 83.85}  & {\bf 85.87} & {\bf 81.93} & {\bf 90.24} & {\bf 53.66} & {\bf 61.94} & {\bf 51.8} & {\bf 40}  & {\bf 24} \\
    \midrule
    WBP-Roth        &  $-$\{word, path\}   & 82.26 ($\pm$0.12) & 84.77 & 79.90 & 89.28 & 50.15 & 57.72 & 49.1 & 38  & {\bf 24}  \\
    BP-Roth         &  $-$\{word, path\}   & 82.03 ($\pm$0.19) & 84.02 & 80.14 & 89.07 & 49.04 & 57.56 & 46.9 & 34  & 18  \\
    WB              &  $-$\{word, path\}   & 82.05 ($\pm$0.19) & 85.42 & 78.95 & 89.18 & 47.21 & 55.42 & 43.9 & 34  & 21  \\
    B               &  $-$\{word, path\}   & 78.54 ($\pm$0.12) & 79.48 & 77.63 & 85.59 & 40.97 & 49.96 & 36.8 & 22  & 9.1 \\
    \bottomrule
	\end{tabular}
	}
	\caption{ Impact of each feature representation.
  ``$-$ word'' indicates the removal of surface and lemma features. ``$-$ cases'' and ``$-$ path'' indicate the removal of the {\it case markers of other dependents} and binary path features, respectively. The task {\it Zero} is equivalent to the cases where the dependency distance $\geq 2$.}
	\label{tbl:model_accuracy}
\end{table*}

\subsection{Experimental details}
\label{ssec:implementation-details}

\paragraph{Dataset}
\label{ssec:dataset}
The experiments were performed on the NTC corpus v1.5, dividing it into commonly used training, development, and test divisions \cite{taira2008japanese}.

\paragraph{Hyperparameters}
We chose the hyperparameters of our models to obtain a maximum score in $F_1$ on the development data.
We select $2,000$ for the dimension of the hidden layers in the feedforward network from $\{256,512,1000,2000,3000\}$,
$2$ for the number of hidden layers from $\{1,2,3,4\}$,
$192$ for the dimension of the hidden unit in GRU from $\{32, 64, 128, 192\}$,
$0.2$ for the dropout rate of GRUs from $\{0.0, 0.1, 0.2, 0.3, 0.5\}$,
and $128$ for the mini-batch size on training from $\{32, 64, 128\}$.

We employed a categorical cross-entropy loss for training, and used Adam with $\beta_1=0.9$, $\beta_2=0.999$, and $\epsilon=1e-08$.
The learning rate for each model was set to $0.0005$. All the model were trained with early stopping method  with a maximum epoch number of 100, and training was terminated after five epochs of unimproved loss on the development data.
The output thresholds for case labels were optimized on the training data.

\paragraph{\bf Initialization}
All the weight matrices in GRU were initialized with random orthonormal matrices.
The word embedding vectors were initialized with 256-dimensional Word2Vec\footnote{https://code.google.com/archive/p/Word2Vec/} vectors trained on the entire
Japanese Wikipedia articles dumped on September 1st, 2016.
We extracted the body texts using WikiExtractor,\footnote{https://github.com/attardi/wikiextractor} and tokenized them using the CaboCha dependency parser v0.68 with JUMAN dictionary. The vectors were trained on lemmatized texts. Adjacent verbal noun and light verb were combined in advance.
Low-frequent words appearing less than five times were replaced by their POS, and we used trained POS vectors for words that were not contained in a lexicon of Wikipedia word vectors in the PAS analysis task.

We used another set of word/POS embedding vectors for lexicalized path embeddings, initialized with 64-dimensional Word2Vec vectors.
The embeddings for dependency directions were randomly initialized. All the pre-trained embedding vectors were fine-tuned in the PAS analysis task.

The hyperparameters for Word2Vec are ``-cbow 1 -window 10 -negative 10 -hs 0 -sample 1e-5 -threads 40 -binary 0 -iter 3 -min-count 10''.

\paragraph{\bf Preprocessing}
We employed a common experimental setting that we had an access to the gold syntactic information, including morpheme segmentations, parts-of-speech, and dependency relations between {\it bunsetsu}s.
However, instead of using the gold syntactic information in NTC, we used the output of CaboCha v0.68 as our input to produce the same word segmentations as in the processed Wikipedia articles.
Note that the training data for the parser contain the same document set as in NTC v1.5, and therefore, the parsing accuracy for NTC was reasonably high.

The binary features appearing less than 10 times in the training data were discarded.
For a path sequence, we skipped a middle part of intermediate tokens and inserted a special symbol in the center of the sequence if the token length exceeded 15.

\subsection{Results}
In the experiment, in order to examine the impact of each feature representation, we prepare arbitrary combinations of word embedding, path embedding, and binary features, and we use them as input to the feedforward network.
For each model name, W, P, and B indicate the use of word embedding, path embedding, and binary features, respectively.
In order to compare the performance of binary features and embedding representations, we also prepare multiple sets of binary features. The evaluations are performed by comparing precision, recall, and $F_1$ on the test set.
These values are the means of five different models trained with the same training data and hyperparameters.

\begin{table*}[!t]
  \centering
	\footnotesize{
	\begin{tabular}{l||l||r|r|r|r||r|r|r|r}
    \toprule
		& & \multicolumn{4}{c||}{Dep} & \multicolumn{4}{c}{Zero}\\
		Model & ALL & ALL & NOM & ACC & DAT & ALL & NOM & ACC & DAT \\
    \midrule
		\multicolumn{9}{c}{On NTC 1.5} \\
		\midrule
    WBP-Shwartz (ens) $-$\{word, path\}    & {\bf 83.85} & {\bf 90.24} & {\bf 91.59} & {\bf 95.29} & 62.61 & {\bf 53.66} & {\bf 56.47} & {\bf 44.7} & {\bf 16} \\
    B                         & 82.02 & 89.11 & 90.45 & 94.61 & 60.91 & 49.59 & 52.73 & 38.3 & 11 \\
		\hline
    \cite{Ouchi2015}-local    & 78.15 & 85.06 & 86.50 & 92.84 & 30.97 & 41.65 & 45.56 & 21.4 & 0.8 \\
		\cite{Ouchi2015}-global   & 79.23 & 86.07 & 88.13 & 92.74 & 38.39 & 44.09 & 48.11 & 24.4 & 4.8 \\
		\cite{ouchi2017}-multi-seq   & 81.42 & 88.17 & 88.75 & 93.68 & {\bf 64.38} & 47.12 &  50.65 & 32.4 & 7.5 \\
    \midrule
    \multicolumn{9}{c}{Subject anaphora resolution on modified NTC, cited from~\cite{Iida2016}} \\
    \midrule
    \cite{Ouchi2015}-global & &&& &&& {\bf 57.3} && \\
    \cite{Iida2015}  & &&& &&& 41.1 && \\
    \cite{Iida2016}  & &&& &&& 52.5 && \\
    \bottomrule
	\end{tabular}
	}
	\caption{Comparisons to previous work in $F_1$}
	\label{tbl:previous_research}
\end{table*}

\paragraph{Impact of feature representations}
\label{sec:feature-impact}

The first row group in Table~\ref{tbl:model_accuracy} shows the impact of the {\it case markers of the other dependents} feature.
Compared with the model using all the binary features, the model without this feature drops by $0.3$ point in $F_1$ for directly dependent arguments ({\it Dep}), and $1.6$ points for indirectly dependent arguments ({\it Zero}).
The result shows that this information significantly improves the prediction in both {\it Dep} and {\it Zero} cases, especially on {\it Zero} argument detection.

The second row group compares the impact of lexicalized path embeddings of two different types.
In our setting, WBP-Roth and WB compete in overall $F_1$, whereas
WBP-Roth is particularly effective for {\it Zero}.
WBP-Shwartz obtains better result compared with WBP-Roth, with further $0.9$ point increase in comparison to the WB model.
Moreover, its performance remains without lexical and path binary features.
The WBP-Shwartz (ens)$-$\{word, path\} model, which is the ensemble of the five WBP-Shwartz$-$\{word, path\} models achieves the best $F_1$ score of $83.85\%$.

To highlight the role of word embedding and path embedding,
we compare B, WB, BP-Roth, and WBP-Roth models on the third row group, without using lexical and path binary features.
When we respectively remove W and P-Roth from WBP-Roth, then the performance decreases by $0.23$ and $0.21$ in $F_1$.
\newcite{Roth2016a} reported that $F_1$ decreased by 10 points or more when path embedding was excluded.
However, in our models, such a big decline occurs when we omit both path and word embeddings.
This result suggests that the word inputs at both ends of the path embedding overlap with the word embedding and the additional effect of the path embedding is rather limited.

\paragraph{Comparison to related work}
\label{sec:implementation-details}

Table~\ref{tbl:previous_research} shows the comparison of $F_1$ with existing research.
First, among our models, the B model that uses only binary features already outperforms the state-of-the-art global model on NTC 1.5~\cite{ouchi2017} in overall $F_1$  with $0.6$ point of improvement.
Moreover, the B model outperforms the global model of \newcite{Ouchi2015} that utilizes the basic feature set hand-crafted by \newcite{imamura2009discriminative} and \newcite{Hayashibe2011} and thus contains almost the same binary features as ours.
These results show that fine feature combinations learned by deep NN contributes significantly to the performance.
The WBP-Shwartz (ens)$-$\{word, path\} model,
which has the highest performance among our models shows a further $1.8$ points improvement in overall $F_1$, which achieves $13$\% error reduction compared with the state-of-the-art grobal model ($81.42$\% of \cite{ouchi2017}-multi-seq).

\newcite{Iida2015} and \newcite{Iida2016} tackled the task of Japanese subject anaphora resolution, which roughly corresponds to the task of detecting {\it Zero} NOM arguments in our task.
Although we cannot directly compare the results with their models due to the different experimental setup, we can indirectly see our model's superiority through the report on \newcite{Iida2016}, wherein the replication of \newcite{Ouchi2015}
showed 57.3\% in $F_1$, whereas \newcite{Iida2015} and \newcite{Iida2016} gave 41.1\% and 52.5\%, respectively.

As a closely related work to ours, \newcite{Shibata2016a} adapted a NN framework to the model of \newcite{Ouchi2015} using a feedforward network for calculating the score of the PAS graph.
However, the model is evaluated on a dataset annotated with a different semantics; therefore, it is difficult to directly compare the results with ours.

Unfortunately, in the present situation, a comprehensive comparison with a broad range of prior studies in this field is quite difficult for many historical reasons (e.g., different datasets, annotation schemata, subtasks, and their own preprocesses or modifications to the dataset).
Creating resources that would enable a fair and comprehensive comparison is one of the important issues in this field.

\section{Conclusion}
\label{sec:conclusion}

This study has argued the importance of designing a sophisticated local model before exploring global solution algorithms in Japanese PAS analysis and empirically demonstrated that a sophisticated local model alone can outperform the state-of-the-art global model with $13$\% error reduction in $F_1$.
This should not be viewed as a matter of local models vs. global models.
Instead, global models are expected to improve the performance by incorporating such a strong local model.

In addition, the local features that we employed in this paper is only a part of those proposed in the literature.
For example, selectional preference between a predicate and arguments is one of the effective information~\cite{sasano2011discriminative,Shibata2016a}, and local models could further improve by combining these extra features.

\section*{Acknowledgments}
This work was partially supported by JSPS KAKENHI Grant Numbers 15H01702 and 15K16045.

\bibliography{ijcnlp2017}
\bibliographystyle{ijcnlp2017}

\end{document}